\documentclass[sigconf,acmengage]{acmart}
\AtBeginDocument{%
  }

\copyrightyear{2026}
\acmYear{2026}
\setcopyright{cc}
\setcctype{by}
\acmConference[HCOMP 2026]{2026 ACM Conference on Human-AI Complementarity and Alignment}{September 27--30, 2026}{Alexandria, VA, USA}
\acmBooktitle{2026 ACM Conference on Human-AI Complementarity and Alignment (HCOMP 2026), September 27--30, 2026, Alexandria, VA, USA}
\acmDOI{10.1145/3834580.3838748}
\acmISBN{979-8-4007-2894-5/2026/09}

\begin{document}

\title{Looks Can be Deceiving: Annotator and Reviewer Performance Across Imagery Sources in\\Crowd-Sourced Aerial Damage Assessment}

\author{Thomas Manzini}
\email{tmanzini@tamu.edu}
\orcid{0009-0002-4937-1948}
\affiliation{
  \institution{Texas A\&M University}
  \city{College Station}
  \state{Texas}
  \country{USA}
}
\author{Priyankari Perali}
\email{perali@umd.edu}
\orcid{0009-0000-2735-1415}
\affiliation{
  \institution{University of Maryland}
  \city{College Park}
  \state{Maryland}
  \country{USA}
}
\author{Raisa Karnik}
\email{raisak@tamu.edu}
\orcid{0009-0008-9184-7499}
\affiliation{
  \institution{Texas A\&M University}
  \city{College Station}
  \state{Texas}
  \country{USA}
}
\author{Stephen Johnson}
\email{sjohnson440@tamu.edu}
\orcid{0009-0005-7610-3182}
\affiliation{
  \institution{Texas A\&M University}
  \city{College Station}
  \state{Texas}
  \country{USA}
}
\author{Robin R. Murphy}
\email{robin.r.murphy@tamu.edu}
\orcid{0000-0003-0774-4312}
\affiliation{
  \institution{Texas A\&M University}
  \city{College Station}
  \state{Texas}
  \country{USA}
}


\begin{abstract}
This paper presents the first known empirical investigation of annotator and reviewer performance across multi-source remotely sensed imagery, evaluating human labeling across drone, crewed aviation, and satellite views.
Because existing aerial imagery datasets rely predominantly on single-source imagery, there is no currently established state of practice for efficiently allocating human labor to curate large-scale, multi-source aerial datasets. This work addresses this limitation by analyzing annotator and reviewer performance within a post-disaster building damage assessment dataset of 9 disasters, where 20,041 buildings in drone, 20,695 buildings in crewed aviation, and 33,392 buildings in satellite imagery were labeled. 
These labels, provided by 187 annotators, were then refined through two successive quality-control stages: a single-reviewer pass followed by a consensus-committee review. Our analysis reveals two findings that raise questions for standard crowd-sourcing practices. First, initial annotations were revised by the final committee at rates that rise steeply from higher- to lower-resolution sources (25.27\% for crewed aviation and 36.95\% for satellite), with the same ordering at every observed workflow stage.
Second, a single individual review reduced but did not resolve this disagreement: after review, the committee still revised 6.85\% of drone, 14.05\% of crewed, and 20.86\% of satellite labels. These observations suggest that, in workflows like this one, uniform review allocation leaves the most residual disagreement in lower-resolution imagery.
Based on this evidence, and consistent with prior work on adaptive task assignment and budget-aware quality control, this paper offers three recommendations for multi-source dataset curation: (1) tailor labeling schemas to each imagery source, (2) prioritize consensus-based adjudication over individual review alone, and (3) preferentially target quality-control effort toward lower-resolution sources.
\end{abstract}

\ccsdesc[300]{Applied computing~Computers in other domains}
\ccsdesc[500]{Computing methodologies~Computer vision}
\ccsdesc[300]{General and reference~Empirical studies}
\ccsdesc[300]{Computing methodologies~Machine learning}

\keywords{Datasets, Quality Control, Crowd-Sourcing, Annotator Performance, Reviewer Performance, Aerial Imagery, Remote Sensing, Human-AI Complementarity, Computer Vision}


\maketitle

\section{Introduction}
\label{sec:intro}

\begin{figure}
    \centering
    \includegraphics[width=\columnwidth]{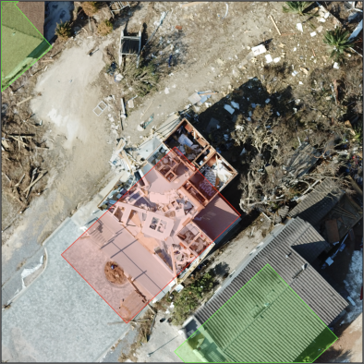}
    \caption{Example image of tile displayed within the annotation tool, LabelBox. All annotators provided annotations through LabelBox, where they labeled each building polygon (shown in green and red) within the image tile.}
    \label{fig:labelbox}
\end{figure}



Remote sensing is increasingly used within operational and societal settings \cite{navalgund2007remote, rindfuss1998linking}. Aerial imagery collected from remote sensors (satellite, crewed aircraft, and drones) offers wide-area spatial views of scenes that can be used for more informed decision-making. In recent years, human-AI systems have been developed to provide rapid assessments of such imagery for societal and scientific applications, such as disaster response \cite{gupta2019creating, manzini2024crasar, scheele2025ladi}, climate monitoring \cite{fu2024remote}, agriculture \cite{mmbando2025harnessing}, and conservation \cite{cavender2022integrating}. However, the fundamental bottleneck in developing these robust models is the human labor required to manually inspect and annotate overhead imagery with prior work relying on crowd-sourced human intelligence spanning drone \cite{manzini2024crasar, rahnemoonfar2023rescuenet, zhu2021msnet, cheng2021dorianet, pi2020convolutional, rahnemoonfar2021floodnet}, crewed \cite{manzini2024crasar, scheele2025ladi, pi2020convolutional}, and satellite \cite{chen2025bright, manzini2024crasar, haitiBRDDataset, lee2022ida, gupta2019creating, fujita2017damage} imagery.

Developing robust human-AI systems for remote sensing requires large-scale, multi-source datasets \cite{ghamisi2019multisource}, yet how varying imagery sources impact the humans tasked with labeling them has been unstudied. Effectively measuring annotator performance requires data collected from the three major remote sensing aerial imagery sources (drones, crewed aircraft, and satellites) labeled under a unified schema and consistent view type (e.g., nadir). While recent efforts have produced multi-source imagery datasets \cite{pi2020convolutional, garioud2023flair, fang2024globally, fawakherji2025deepflood}, they fail to meet these criteria: they lack one of the three imagery sources, utilize disjointed schemas across sources, or mix view types (e.g., oblique vs. nadir). Although DeepFlood \cite{fawakherji2025deepflood} provides all three sources, the satellite component consists of non-visual data not labeled by annotators. This leaves CRASAR-U-DROIDs \cite{manzini2024crasar} as the only known dataset that possesses all three imagery sources independently labeled under a unified schema and view type. As a result, the annotation and review data collected during the development of CRASAR-U-DROIDs becomes the necessary subject of this work.


Curating such large-scale multi-source datasets to support CV/ML efforts is resource-intensive, where reviewing every annotated label for quality control is often impractical. Along with crowd-sourcing, prior efforts have employed strategies to provide a level of quality control, such as random sampling the annotations for review \cite{gupta2019creating} or by an exhaustive review of all annotations, by either one reviewer \cite{rahnemoonfar2023rescuenet, rahnemoonfar2021floodnet} or a committee of reviewers \cite{manzini2024crasar}. 
Consequently, the two-stage review pipeline utilized during the curation of the CRASAR-U-DROIDs dataset provides a rare empirical testbed. Rather than treating this workflow merely as a rigid curation mechanism, it allows for successive quality-control stages, a single-reviewer verification pass followed by a consensus-committee review, to be observed and quantified at scale on the same labels.

Recent findings highlight that curating multi-source imagery cannot rely on a direct, 1:1 mapping of labels across sources, as different sources
(e.g., drone vs. satellite) alter the perceived content of the imagery \cite{manzini2025now}. Consequently, each imagery source requires its own dedicated human annotation and review pipeline. Despite this massive requirement for independent human labor, there is no prior work investigating how annotator and reviewer performance actually differ across these sources. Without this understanding, there is a limited understanding of the human phenomena that drive the curation of multi-source datasets. This drives the motivating question for this work: \emph{How can annotator and reviewer efforts be targeted for efficient curation of high-quality large-scale multi-source aerial imagery datasets?}

Guided by this research question, this work presents the first and largest known multi-source analysis of annotator and reviewer performance. Utilizing the CRASAR-U-DROIDs \cite{manzini2024crasar} dataset, this work analyzes annotator and reviewer performance across 20,041 drone-derived, 20,695 crewed-derived, and 33,392 satellite-derived building labels observed at three workflow stages: initial annotation, initial review, and committee review. This analysis represents the core contribution of this work, yielding two findings: initial annotations were revised by the final committee at rates that surge as source resolution coarsens (25.27\% for crewed aviation and 36.95\% for satellite), and a single individual review left substantial residual revision behind (6.85\% for drone, 14.05\% for crewed, and 20.86\% for satellite). Based on this evidence, this paper suggests three practices for multi-source dataset curation: (1) tailor labeling schemas to each imagery source, (2) prioritize consensus-based adjudication over individual review alone, and (3) preferentially target review capacity toward lower-resolution sources.

\section{Background \& Related Work}
\label{sec:related_work}

Curating robust datasets for remote sensing requires a deep understanding of human computation, yet annotator and reviewer performance across multi-source imagery remains largely unstudied. To contextualize this gap, this section first outlines the growing reliance on crowd-sourcing to overcome the human labor bottleneck in remote sensing (Section \ref{subsec:crowdsourcing_rs}) and the parallel shift toward multi-source data fusion (Section \ref{subsec:multi_source_rs}). The post-disaster damage assessment is examined as an operational case study to highlight the disjointed quality control strategies currently used in aerial imagery curation (Section \ref{subsec:quality_control_aerial}). Finally, prior analyses of annotator performance are reviewed and the necessity of evaluating consensus-based review mechanisms across diverse sensing platforms is established (Section \ref{subsec:prior_analysis_annotator_reviewer}).


\subsection{Crowd-sourcing and Human-AI Systems in Remote Sensing}
\label{subsec:crowdsourcing_rs}

The development of robust human-AI systems in remote sensing is fundamentally constrained by the availability of labeled data. While modern remote sensors collect vast volumes of aerial imagery, transforming this raw data into actionable ground truth requires massive amounts of manual human annotation. This requirement creates a severe labor bottleneck, slowing the training and deployment of downstream computer vision models \cite{xia2018dota}.

To overcome this bottleneck, the remote sensing community increasingly relies on human computation and crowd-sourcing \cite{fritz2017role, huang2024crowdsourcing, saralioglu2020crowdsourcing}. By distributing annotation tasks to large pools of non-expert workers or volunteers, researchers can rapidly scale the labeling of aerial datasets for tasks such as land cover classification \cite{fritz2012geo} and large-scale spatial feature tagging \cite{lin2014crowdsourcing}.

However, overhead aerial imagery introduces severe perceptual challenges not present in standard, ground-level computer vision tasks. Annotators must interpret nadir (top-down) 
perspectives, arbitrary orientations, variable ground sample distances, and sensor-specific artifacts \cite{xia2018dota}. These observational complexities make crowd-sourced aerial annotations inherently prone to noise and high variance. Consequently, deploying crowd-sourced labor for aerial imagery necessitates rigorous quality control and review workflows to ensure the resulting datasets are reliable enough for model training \cite{fritz2017role}.

\subsection{Multi-Source Remote Sensing}
\label{subsec:multi_source_rs}

To build more robust human-AI systems, the remote sensing community is increasingly transitioning toward multi-source data fusion, integrating observations from satellites, crewed aircraft, and drones \cite{ghamisi2019multisource}. By leveraging multiple sensing platforms, downstream models can overcome the temporal and spatial limitations inherent to any single source. This operational shift has driven the recent curation of large-scale multi-source datasets. For example, the FLAIR dataset incorporates multi-source optical imagery for country-scale land cover semantic segmentation \cite{garioud2023flair}, while other efforts have generated multi-source datasets for coseismic landslide mapping \cite{fang2024globally} and inundated vegetation segmentation \cite{fawakherji2025deepflood}.

While these efforts advance the availability of diverse remote sensing data, they lack the necessary structure to evaluate human annotator performance across differing imagery sources. Existing multi-source datasets typically fail to meet three strict criteria required for comparative annotator analysis: they lack visual data across all three primary platforms, they utilize disjointed labeling schemas across sources, or they mix fundamentally different view types, such as oblique versus nadir perspectives. For instance, while DeepFlood \cite{fawakherji2025deepflood} provides multi-source data, its satellite component consists of non-visual data that human annotators do not manually label. Consequently, there is a critical absence of multi-source datasets featuring independent human annotations collected under a unified schema and consistent view type. This structural deficit obscures how differing imagery sources inherently impact the human labor tasked with curating them.

\subsection{Quality Control in Aerial Imagery Labeling}
\label{subsec:quality_control_aerial}

Post-disaster building damage assessment provides one of the most mature operational testbeds for examining crowd-sourced annotation workflows in remote sensing. Twelve post-disaster building damage assessment datasets containing drone, crewed aviation, or satellite imagery were identified from prior literature \cite{manzini2024crasar, rahnemoonfar2023rescuenet, zhu2021msnet, cheng2021dorianet, rahnemoonfar2021floodnet, scheele2025ladi, pi2020convolutional, chen2025bright, haitiBRDDataset, lee2022ida, gupta2019creating, fujita2017damage}. Crucially, ten of these are strictly single-source datasets \cite{rahnemoonfar2023rescuenet, rahnemoonfar2021floodnet, zhu2021msnet, cheng2021dorianet, scheele2025ladi, gupta2019creating, haitiBRDDataset, fujita2017damage, lee2022ida, chen2025bright}. Only two contain multi-source imagery: Volan v.2018 \cite{pi2020convolutional} (drone and crewed) and CRASAR-U-DROIDs \cite{manzini2024crasar} (drone, crewed, and satellite).

In all known aerial imagery damage assessment datasets, human annotators perform the initial labeling. Annotator populations typically belong to three distinct groups: in-house research teams \cite{haitiBRDDataset, lee2022ida, pi2020convolutional}, recruited domain experts or trained volunteers \cite{chen2025bright, zhu2021msnet, scheele2025ladi}, and student or crowd-worker populations \cite{manzini2024crasar}.

Despite this reliance on human labor, there is no standardized consensus on quality control and review strategies. Prior curation efforts typically employ at least one reviewer to inspect labels, but the exact mechanism varies widely. For example, the xBD dataset \cite{gupta2019creating} utilized a two-tiered system where all labels received a basic check for obvious errors, followed by an expert review of a 4\% random sample. Other datasets mandate an exhaustive review of all annotations by earth observation experts \cite{chen2025bright}. Several datasets utilize "in-house" reviewers \cite{haitiBRDDataset, lee2022ida, pi2020convolutional, rahnemoonfar2021floodnet, rahnemoonfar2023rescuenet, manzini2024crasar}, often enforcing a single-reviewer pass where annotations are iteratively rejected or approved \cite{rahnemoonfar2021floodnet, rahnemoonfar2023rescuenet}. The LADI v2 dataset \cite{scheele2025ladi} instead enforced quality control dynamically through inter-annotator agreement, requiring at least three annotators per image. Finally, CRASAR-U-DROIDs \cite{manzini2024crasar} utilized a hybrid 2-stage review process consisting of a single-reviewer pass followed by a consensus-committee review.

This disjointed landscape highlights a critical gap in the literature: there is no documented analysis of how effectively these human reviewers perform, particularly across multi-source aerial imagery. Without understanding the failure rates of single-reviewer versus consensus mechanisms across different sensing platforms, future efforts to efficiently expand and curate operational multi-source datasets will remain hindered by unverified assumptions about human labor allocation.

\begin{figure*}[!ht]
    \centering
    \includegraphics[width=\textwidth]{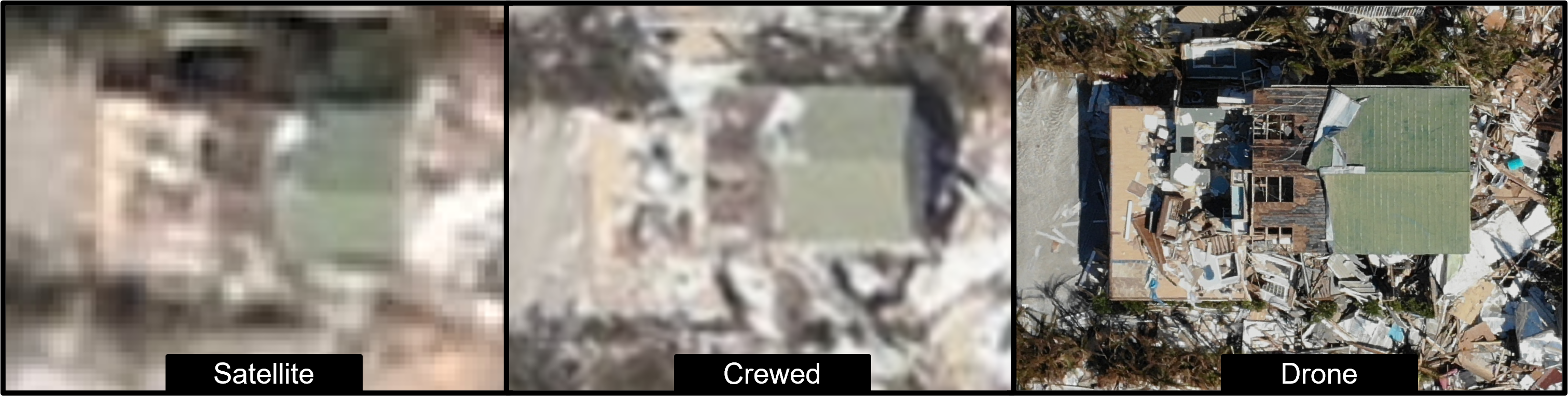}
    \caption{Example of building across three sources of imagery from the CRASAR-U-DROIDs dataset\cite{manzini2024crasar} following Hurricane Ian. [Left] View of building from Maxar satellite imagery (30cm/px). [Middle] View of building from NOAA crewed aerial imagery (15cm/px). [Right] View of building from FL-UAS1 uncrewed (Drone) aerial imagery (3.35cm/px). Notice changes in resolution and appearance across sources.}
    \label{fig:cross_view_building}
\end{figure*}

\subsection{Prior Analyses of Annotator Performance}
\label{subsec:prior_analysis_annotator_reviewer}

The broader computer vision and natural language communities have increasingly recognized the necessity of rigorous dataset annotation quality management, demonstrating that relying on unverified crowd-sourced labels severely degrades model performance \cite{klie2024analyzing, nassar2019assessing}. Prior studies emphasize that efficient, statistical quality estimation and inter-annotator agreement metrics are critical for identifying noise in large-scale datasets \cite{klie2024efficient, nassar2019assessing}. However, these generalized frameworks rarely account for the unique spatial and observational complexities inherent to remote sensing.

Within the remote sensing domain, three relevant prior analyses have investigated annotator performance \cite{wang2023accurate, wang2024well, blushtein2025performance}. \citeauthor{wang2023accurate} \cite{wang2023accurate} evaluated confidence-weighting mechanisms for crowd-sourced annotations to reduce the need for expert intervention. Similarly, \citeauthor{wang2024well} \cite{wang2024well} analyzed 50 annotators labeling land cover in satellite imagery, finding that dedicated training improves accuracy and that inter-annotator agreement is a strong indicator of reliability. Most relevantly, \citeauthor{blushtein2025performance} \cite{blushtein2025performance} analyzed the performance of 25 annotators on object detection tasks across four annotation strategies. Crucially, their analysis found that a majority-vote annotation strategy outperformed independent expert review efforts. 

A separate line of quality-control research treats annotation as an allocation problem: adaptive task assignment routes items and workers by estimated difficulty and skill \cite{ho2013adaptive}, selective relabeling asks when an existing label merits another look \cite{sheng2008get, lin2014re}, and budget-aware designs optimize labeling and review under a fixed budget \cite{karger2014budget, dai2010decision}. Structured adjudication of expert disagreement has been studied directly \cite{schaekermann2019understanding}. However, these mechanisms have rarely been examined in production-scale, multi-source remote sensing workflows, which is the setting this work observes.

While the remote sensing analyses above suggest that consensus mechanisms can outperform individual reviews, they are limited to single-source imagery. It remains unstudied whether these findings hold across large-scale, multi-source aerial imagery datasets, and how annotator and reviewer performance shifts across sensing platforms. This work addresses this gap empirically.
 
\section{Approach}
\label{sec:approach}

This analysis considers 20,041 drone-, 20,695 crewed-, and 33,392 satellite-derived damage labels provided within the CRASAR-U-DROIDs dataset \cite{manzini2024crasar}. Due to its unique feature of containing coincident buildings across three imagery sources (drone, crewed, and satellite), the dataset was selected to be the most appropriate for analyzing revision rates across imagery sources and review stages. This section will detail the data analyzed (Section \ref{subsec:data}), the annotator population (Section \ref{subsec:annotator_population}), and the labeling workflow (Section \ref{subsec:labeling_workflow}) used for the dataset's curation. 

\subsection{Data}
\label{subsec:data}

The selection of the CRASAR-U-DROIDs dataset \cite{manzini2024crasar} was due to its multi-source imagery, comprising drone, crewed, and satellite (an example is shown in Figure \ref{fig:cross_view_building}), and large scale, providing 74,128 building damage labels. This dataset faithfully represents drone-, crewed-, and satellite-based imagery, as there is a stratification of the image resolutions collected in practice as compared to theoretical resolutions \cite{manzini2025now, lozano2023data}.

With the data sourced from CRASAR-U-DROIDs, the annotation and review data were obtained from a 2-stage labeling workflow, providing an opportunity to analyze annotation and review quality for each source of imagery, which is more extensive than any other prior dataset. Annotator and reviewer statistics and performance were derived from a combination of annotations and metrics stored within LabelBox \cite{Labelbox2024} and the dataset. All labels analyzed were initially provided by annotators via the annotation tool, LabelBox \cite{Labelbox2024}. Only building damage labels that were present in all review stages were considered, and any ``manual"  and ``bulk" buildings were discarded \cite{manzini2024crasar}.

The building damage assessment labels were provided for post-disaster imagery sourced by either drone, crewed, or satellite, with ground sampling distances at approximately 3cm/px, 15cm/px, and 30cm/px, respectively. The drone imagery was sourced from the Center for Robot-Assisted Search and Rescue (CRASAR) and collected from nine federally declared disasters, comprising six hurricanes (Ian, Idalia, Ida, Michael, Harvey, Laura), the Mayfield Tornado, the Kilauea Volcano Eruption, and the Musset Bayou Fire. The crewed imagery was sourced from NOAA\cite{noaa} and was collected from Hurricanes Ian, Ida, Idalia, Laura, Harvey, and Michael. The satellite imagery was sourced from the MAXAR Open data portal \cite{maxar15} and was collected from Hurricanes Ian, Idalia, Harvey, and Michael. Of these, the coincident labels analyzed in this work span eight of the nine events for drone imagery, six for crewed aviation, and three for satellite\footnote{Drone imagery from Hurricane Harvey, and Satellite imagery from Hurricane Idalia was annotated in ``bulk" by the dataset curation team instead of the annotator pool, was thus excluded from this analysis}.

\begin{figure*}[!h]
    \centering
    \includegraphics[width=0.85\textwidth]{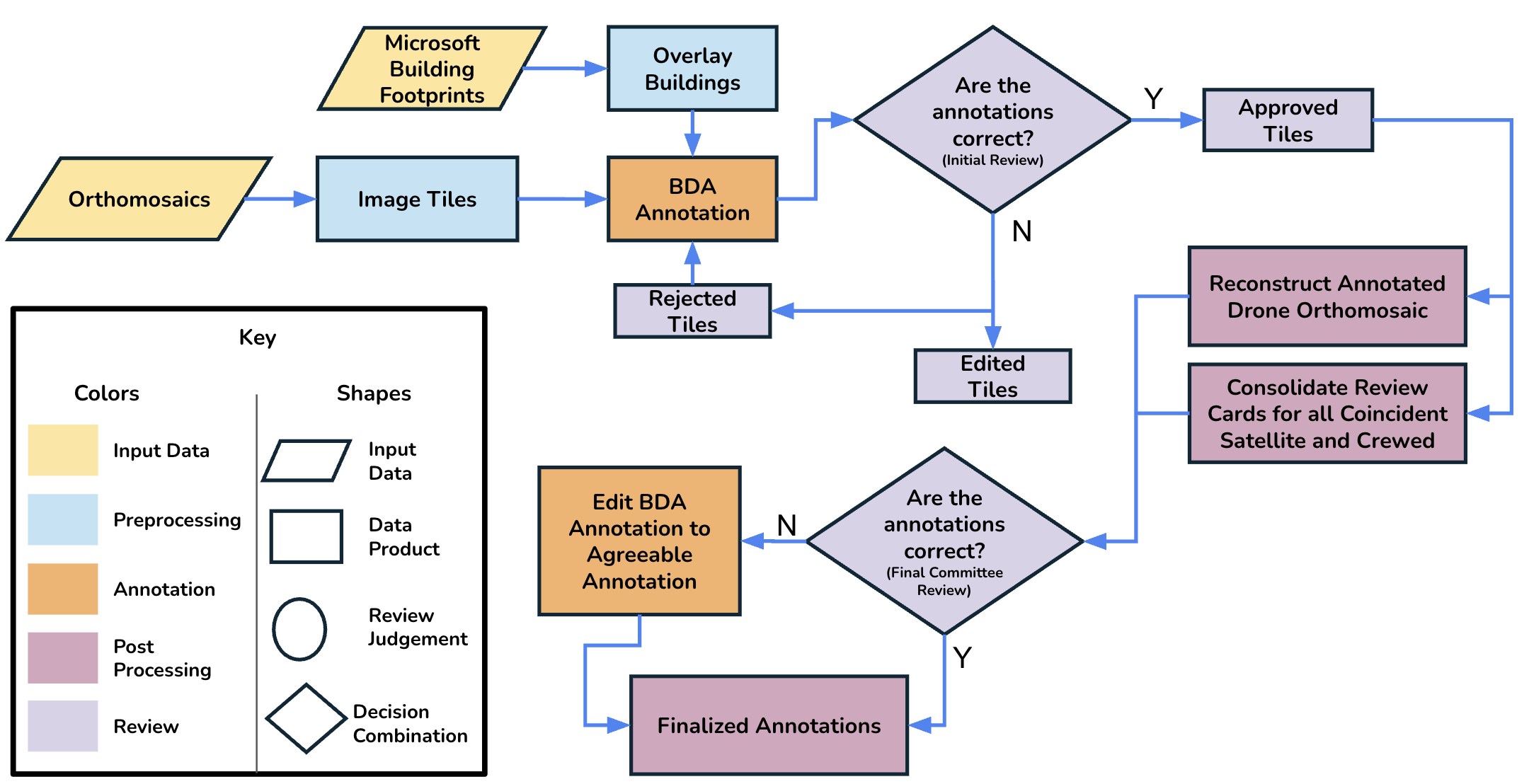}
    \caption{Annotation and review workflow used to annotate 43,223 Images of 74,128 building views across 3 imagery sources (Drone, Crewed, Satellite). }
    \label{fig:labeling_workflow}
\end{figure*}

\subsection{Annotator Population}
\label{subsec:annotator_population}

187 total annotators participated in the collection of building damage labels and are considered in this analysis. Of the 187 total annotators, 172 were high-school students, 7 were middle-school students, and 8 were undergraduate or graduate students associated directly with the research effort who also served as reviewers. 

The 179 high-school and middle-school students who voluntarily participated in the annotation effort did so through STEM outreach events organized to teach students about how machine learning systems are trained and used. Students from 5 high schools and 1 middle school participated in the effort. Students participating were compensated via educational guest lectures on machine learning and ``community service hours," which were reported to their instructors at the end of the semesters in which students participated.

Among the total 187 unique annotators, 55 labeled drone imagery, 67 labeled crewed aircraft imagery, and 93 labeled satellite imagery. 22 annotators labeled more than one source of imagery, driven by expressed interest by the annotators, or through participation in the research team. 

It is worth noting that this annotator population differs from the traditional populations found in crowdworking platforms such as Amazon's Mechanical Turk \cite{mturk}. The population of Mechanical Turk Workers is well studied \cite{ross2010crowdworkers, ipeirotis2010demographics, difallah2018demographics} and differs from the population considered in this work primarily in age and education. Prior work has found that 54\% of Mechanical Turk workers are between 21-35 years old \cite{levay2016demographic}, with 55\% of Mechanical Turk workers not having a bachelor's degree \cite{levay2016demographic}. This represents the primary difference compared to the population considered in this work, of whom 96\% have not earned a bachelor's degree, and under 21 years of age.

\subsection{Labeling Workflow}
\label{subsec:labeling_workflow}
The curation of the data in this work employed a 2-stage review process for 43,223 images of 74,128 buildings across 3 imagery sources (drone, crewed, satellite), annotated by 187 annotators for building damage assessment according to the Joint Damage Scale (JDS)\cite{gupta2019creating}. This process is shown in Figure \ref{fig:labeling_workflow}. The labeling workflow consisted of five primary components: input data, preprocessing, annotation, post-processing, and review. Differences between this population and the population traditionally used in crowd-working tasks \cite{ross2010crowdworkers, levay2016demographic, saravanos2021hidden} suggested that additional oversight was needed to ensure annotation quality.

The input data consisted of orthomosaics and building polygons. An orthomosaic refers to a large spatial area map, where each pixel has a longitude and a latitude. The multi-source imagery, detailed in Section \ref{subsec:data}, was provided as orthomosaics. The building polygons were extracted from the Microsoft Building Footprints dataset \cite{MicrosoftBuildingFootprints}. 

The preprocessing steps consisted of image tiling and overlaying building polygons. Drone orthomosaics were tiled into 2048$\times$2048-sized tiles, and crewed and satellite orthomosaics were tiled into 256$\times$256-sized tiles. The difference kept the ground area covered per tile within a comparable range. The orthomosaics were tiled with a 5\%
overlap and overlaid with building polygons, resulting in 43,223 image tiles. 

As the image tiling process was grid-based, buildings were frequently divided across multiple image tiles, as shown in Figure \ref{fig:labelbox}. This division of buildings across multiple tiles resulted in the 74,128 buildings being presented to annotators in the form of 153,178 unique sub-polygons. These 153,178 sub-polygons would later be recombined for the 74,128 building polygons.

There were three annotation opportunities: ``Initial Annotation," ``Initial Review", and ``Final Committee Review." During Initial Annotation, the 153,178 sub-polygons shown in the 43,223 image tiles representing the 74,128 buildings were annotated by 187 annotators through the annotation tool, LabelBox \cite{Labelbox2024}. Initial Annotation began with an approximately 7-minute instructional session detailing how to annotate the building sub-polygons according to the JDS and the mechanics of how to use the LabelBox annotation tool \cite{gupta2019creating}.
The same instructional session was provided to the annotators for all three sources of imagery, with only the examples used differing depending on the source of imagery. 
Immediately after this session, annotators began providing labels. This consisted of a single annotator viewing an image tile and its associated sub-polygons, and providing labels for all the sub-polygons in that image tile. For approximately 45 minutes after this instructional session, the organizers provided real-time feedback on the usage of the LabelBox tool and corrected misconceptions about the application of the JDS. Once this period had elapsed, annotators were permitted to provide annotations without supervision. 

The Initial Review consisted of a single reviewer inspecting the labels provided within LabelBox. These reviewers would either reject, edit, or approve the annotations. If the reviewer rejected a tile, it would be requeued and sent to a different annotator. Reviewers could also choose to manually correct tiles. This review stage was completed once all image tiles were approved. Across the corpus, this stage was carried out by a pool of three to four distinct reviewers per source, including two research-team members in each; each tile received one reviewer.

Postprocessing followed in order to prepare the annotations for the Final Committee Review. All annotations provided by the 187 annotators were postprocessed to recombine and overlay the annotations over the original imagery. Postprocessing was the recombination of the 153,178 sub-polygons into the 74,128 building polygons. This was performed by tabulating all of the labels for the sub-polygons and coalescing their labels. In cases where the labels for all sub-polygons agreed, the label was simply that one label. In cases where there was disagreement, the highest damage label was selected, which biases pre-committee building labels toward higher damage by construction.

The Final Committee Review followed and consisted of a committee of reviewers inspecting the reconstructed imagery and labels and correcting any errors. For drone imagery, the fused annotations were overlaid over the original orthomosaics. For crewed and satellite imagery, the fused annotations were overlaid on the original imagery, but provided within review cards\footnote{An example Review Card is shown in Appendix \ref{appendix:review_cards}.}.
This stage of review was complete once all annotations were approved by the review committee, thus finalizing the annotations.
\section{Analysis}
\label{sec:analysis}

The analysis conducted in this work seeks to answer the motivating question introduced in Section \ref{sec:intro}. This analysis considers both the performance characteristics of the annotators within the LabelBox tool and their agreement with the recombined building-level reference labels (Section \ref{subsec:limitations}).

\begin{figure}
    \centering
    \includegraphics[width=\columnwidth]{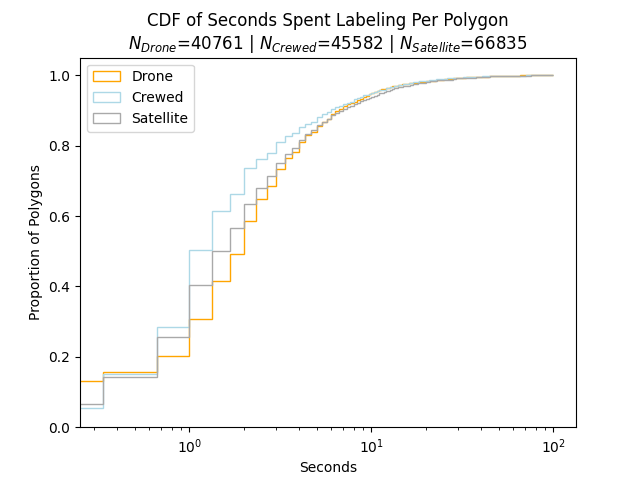}

    \vspace{0.0cm} 
    
    \centering
    \begin{tabular}{|l|lllll|}
    \hline
    Percentile    & \multicolumn{1}{l|}{0.5} & \multicolumn{1}{l|}{0.75} & \multicolumn{1}{l|}{0.90} & \multicolumn{1}{l|}{0.95} & \multicolumn{1}{l|}{0.99} \\ \hline
    Drone & \begin{small}2.0 \end{small}                    & \begin{small}3.5 \end{small}                    & \begin{small}6.75 \end{small}                    & \begin{small}10.5 \end{small}                    & \begin{small}28.0 \end{small} \\ \hline
    Crewed & \begin{small}1.29 \end{small}                    & \begin{small}2.5\end{small}                    & \begin{small}6.13\end{small}                    & \begin{small}10.36\end{small}                    & \begin{small}26.5\end{small}\\ \hline
    Satellite & \begin{small}1.66 \end{small}                    & \begin{small}3.33\end{small}                    & \begin{small}7.04\end{small}                    & \begin{small}11.95\end{small}                    & \begin{small}30.92\end{small} \\ \hline
    \end{tabular}
    \caption{CDF of seconds annotators spent labeling per polygon across drone, crewed \& satellite imagery. Observe that median per-polygon annotation times are similar across imagery sources. Note the x-axis log scale. Percentiles are shown in seconds.}   
    \label{fig:cdf_annotators}
\end{figure}

\begin{figure*}
    \centering
    \includegraphics[width=\textwidth]{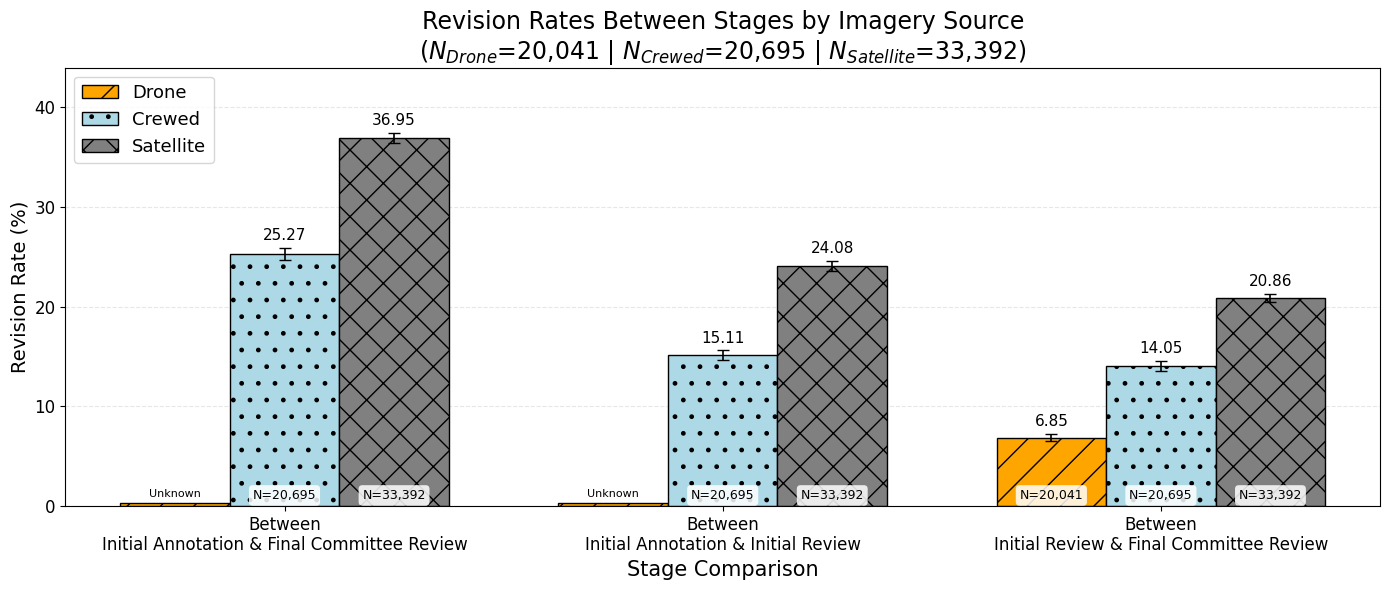}
    \caption{Label revision rates between workflow stages (Initial Annotation, Initial Review, and Final Committee Review). Whiskers show 95\% confidence intervals under a building-label independence assumption. Rates that could not be computed (pre-review drone labels were not preserved) are marked ``unknown". Note the higher revision rates against Initial Annotation and for lower-resolution sources (e.g., Satellite).}
    \label{fig:revision_rates}
\end{figure*}

\subsection{Annotation Timing}

As discussed, within the LabelBox annotation tool, 187 annotators provided labels for 74,128 buildings, which were represented through 153,178 building sub-polygons displayed within 43,223 image tiles. At the median, Initial Annotation took 2 seconds per drone tile, 4 seconds per crewed tile, and 8 seconds per satellite tile. It should be noted that the LabelBox tool records annotation time as an integer number of seconds. However, the differing image resolutions meant that a differing count of buildings was shown in each tile depending on its pixel dimensions.

To measure the time spent annotating per building sub-polygon within a given tile, the integer number of seconds spent annotating is divided by the number of building sub-polygons within the tile. When doing this, the trend observed at the tile level disappears. At the median, drone imagery takes 2.0 seconds per building polygon, crewed imagery takes 1.29 seconds, and satellite imagery takes 1.66 seconds\footnote{The statistical significance of the differences between sources cannot be reliably computed due to the integer nature of the timing information recorded by LabelBox.}. The complete distribution for drone, crewed, and satellite labeling time per building sub-polygon is shown in Figure \ref{fig:cdf_annotators}. This finding suggests that there is little difference in the time spent annotating sub-polygons across resolutions.

\subsection{Label Revision Rates}
\label{sec:revision_rates}
This work reports \emph{revision rates}: the fraction of buildings present at both compared stages whose damage label differs. The Final Committee Review labels serve as the workflow's determinative reference, not independently verified ground truth (Section~\ref{subsec:limitations}). As shown in Figure \ref{fig:revision_rates}, revision rates rose from drone to crewed to satellite imagery in every stage comparison, ranging from 6.85\% to 36.95\%; drone comparisons involving Initial Annotation could not be computed.

The error bars in Figure \ref{fig:revision_rates} give the 95\% Wilson confidence intervals, treating buildings as independent samples, reflecting how the labels were elicited, since each building was presented to annotators and reviewers as its own labeling decision. Labels do share context (a single annotator labeled many buildings, different disasters impact different buildings, and satellite buildings recur across views), so these intervals capture sampling variability under a building-level independence assumption and may understate uncertainty from such shared factors (Section \ref{subsec:limitations}); annotator effects in particular cannot be modeled separately from source effects, as annotator cohorts across sources (Section \ref{subsec:annotator_population}) were not consistent. Revision rates differed significantly between sources within every stage pair. After individual review, the odds of a committee revision for a satellite label were $3.6\times$ those of a drone label (odds ratio 3.6, 95\% CI [3.4, 3.8]) and $1.6\times$ those of a crewed label (95\% CI [1.5, 1.7]). Moving beyond confidence intervals, it is argued that Stuart-Maxwell marginal-homogeneity tests are the most appropriate in this context because successive stages relabel the same buildings. As a result, distributional shifts between stages are tested with Stuart--Maxwell marginal-homogeneity tests and all found to be $p < .001$. Further, statistical significance was also observed via chi-squared tests, as all $p < .001$ with Cram\'er's $V$ = 0.11--0.16.

When an individual review changed an annotator's label, the committee endorsed the reviewer's label in 83.0\% (crewed) and 77.2\% (satellite) of cases, reverting to the original annotation in only 8.7\% and 10.4\%. Most committee revisions instead fell on buildings the review had left unchanged (13.5\% and 20.3\% of which were revised). In this workflow, individual reviewers primarily diverged from the committee by not intervening rather than by intervening differently.

\subsection{Label Revision Directionality}

\begin{figure*}
    \centering
    \includegraphics[width=\textwidth]{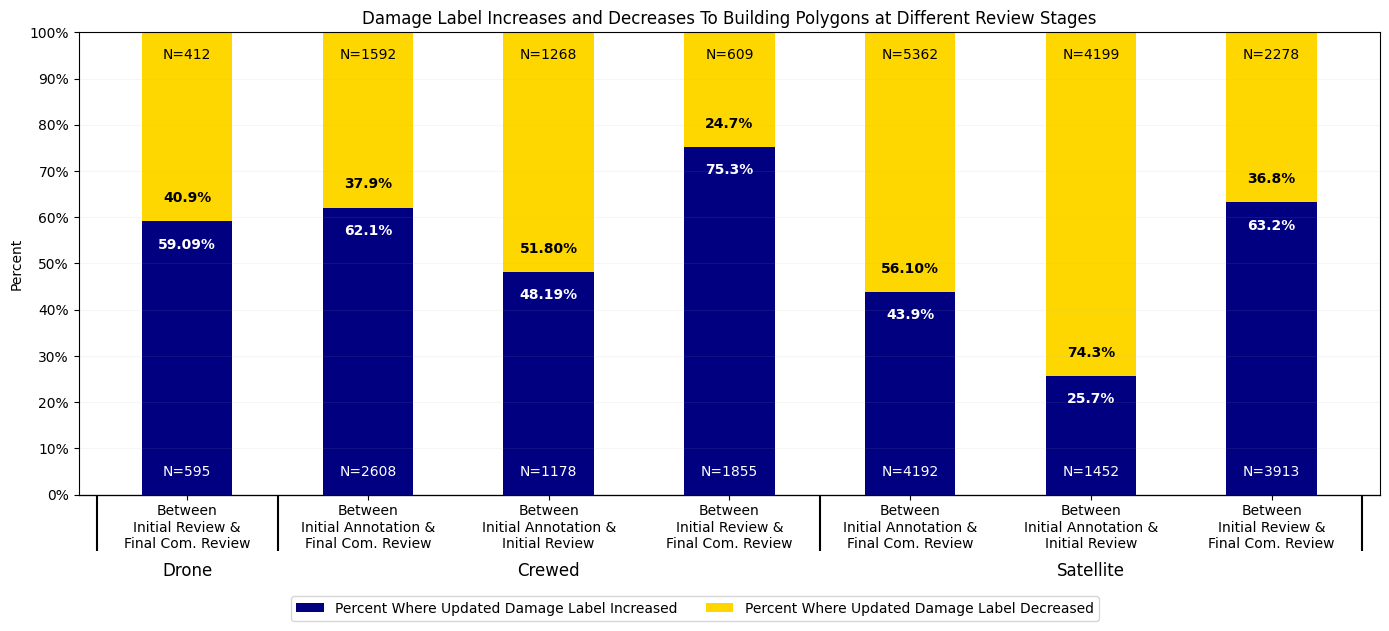}
    \caption{Direction of label revisions between workflow stages. Under- or over-estimation of damage is measured by the changes made to the building labels at each step of review. Updates to building polygon labels that resulted in a change to or from a label that is not associated with damage (obscured, un-classified) were not included.}
    \label{fig:revision_directionality}
\end{figure*}

Figure \ref{fig:revision_directionality} shows the direction of revisions at each observed stage pair. Four of the seven comparisons were dominated by increases: relative to the labels under review, the Final Committee Review predominantly raised damage severity in all three sources (59.1\% of its direction-changing revisions for drone, 75.3\% for crewed, and 63.2\% for satellite). The Initial Review stage differed by source: near-balanced for crewed (48.2\% increases) and predominantly lowering for satellite (74.3\% decreases), driven by demotions of minor- and major-damage annotations toward no damage.

Overall, the determinative labels sat above the crewed Initial Annotations (62.1\% of directional revisions increased severity) but below the satellite Initial Annotations (56.1\% decreased): annotators under-reported damage in crewed imagery and over-reported it in satellite imagery, so the direction of annotator bias is itself source-dependent. The mechanisms behind these source-dependent directions cannot be isolated in this observational design (Section \ref{subsec:limitations}).

\section{Discussion}
\label{sec:discussion}

The analysis in Section \ref{sec:analysis} finds that revision rates significantly differed between imagery sources and review stages, higher for lower-resolution sources, and largest against Initial Annotation, implicating three areas of consideration for curating multi-source aerial imagery datasets. This section first discusses limitations (Section \ref{subsec:limitations}), the three implications with recommendations for improvements (Section \ref{subsec:implications}), and ethical considerations (Section \ref{subsec:ethics}).

\subsection{Limitations}
\label{subsec:limitations}

There are four limitations of this work, based on the availability of imagery, damage labels, and their associated annotator and reviewer data. 

First, this work inherently evaluates varying imagery sources rather than varying sources and ground sample distances. While evaluating synthetically downsampled or upsampled imagery would isolate the variable of pixel density, it would fail to capture the atmospheric interference, distinct sensor artifacts, and off-nadir angle distortions unique to the imagery source. Therefore, the dynamics observed here represent the compounded effects of true multi-source imagery, rather than isolated resolution degradation. Beyond sensing differences, the sources also differ in disaster coverage and the annotator cohorts.  As a result, the statistics reported here reflect these factors jointly, and no single cause, including resolution, can be isolated.

Second, this work treats the labels arrived at following the final committee review as the determinative label, and reports revision rates against them; they are not independently verified ground truth. This is notably different from labels generated by an expert inspector at the building site on the ground \cite{fema_pda_2025}, and prior work has found differences in the distribution of labels generated from different sources \cite{manzini2025now}. Consequently, a revision measures divergence between workflow stages, not error against an external standard.

Third, as mentioned in Section \ref{sec:analysis}, the initial drone labels, prior to their correction by reviewers, are not preserved in LabelBox. This was only recognized after all initial reviews had been completed. The data was later manually preserved for the satellite and crewed aircraft imagery annotations, which were labeled and reviewed afterward.

Fourth, the imagery considered in this dataset, and the annotation task itself, is limited to hurricanes, a volcanic eruption, a tornado, and a wildfire. This means that it is unknown if the dynamics observed in this work will persist in other imagery of different disaster types or annotation tasks.

\subsection{Implications and Recommendations for Annotation and Reviews Across Imagery Sources}
\label{subsec:implications}

The findings of the analysis presented in Section \ref{sec:analysis} imply three items for future data annotation and review efforts for computer vision and machine learning datasets with potential improvements. 

The first implication concerns allocation: revision rates in this workflow varied widely across sources, so uniformly allocating a fixed amount of review effort left the largest residual disagreement in the lower-resolution sources. This evidence leads to the recommendation that review capacity is better targeted toward the sources that accumulate the most revisions.

Second, if label revisions concentrate in lower-resolution sources and go uncorrected, models trained on such data may inherit source-dependent biases; this risk is anticipated here but not measured, and benchmarking it is left to future work. To reduce it, labeling schemas should anticipate each imagery source, since the guidance annotators need appears to differ by source.

Third, individual review alone did not align labels with the final committee outcome. For crewed and satellite imagery, the committee revised 14.05\% and 20.86\% of post-review labels, respectively, fewer than the 25.27\% and 36.95\% revisions of the initial annotations, so individual review moved labels toward the eventual committee outcome. However, a substantial residual remained, concentrated in buildings the review had left unchanged (Section \ref{sec:revision_rates}). Because the committee outcome is itself the reference (Section \ref{subsec:limitations}), these residuals measure how far a single review leaves labels from the workflow's determinative label, not reviewer error against ground truth.

The remaining disagreement after Initial Review underscores the limits of a single-review stage process. Individual reviewers assign labels without the opportunity to discuss uncertainty when faced with ambiguous imagery, whereas a consensus committee can deliberate on complex edge cases \cite{surowiecki2005wisdom, sheng2008get}; deliberation is also precisely why the committee outcome serves as the workflow's determinative reference rather than an independent measurement.

\subsection{Ethical Implications for Label Collection}
\label{subsec:ethics}
Crowd-sourcing labels can cause potential negative implications for downstream operations. This work raises three key ethical considerations when crowd-sourcing labels for multi-source datasets: label variation can induce bias and de-calibrate models, constraints on downstream uses, and the need for expert review of annotations to mitigate errors introduced by untrained annotators. 

Due to a lack of subject-matter expertise in crowd-sourced annotators, labels created by crowd workers may have a large spread of variance. This variance can introduce bias into downstream training and cause de-calibration, restricting potential deployments for trained models and degrading their reliability for operational uses. These concerns motivate expert review: consolidating and examining labels lets experts catch misclassifications and dampen crowd-induced variance, and adjudicating rather than authoring labels can limit the bias experts introduce.
\section{Conclusion}
\label{sec:conclusion}

This work contributes the first and largest known empirical analysis of human annotator performance across multi-source remotely sensed imagery. By evaluating 74,128 annotated buildings labeled by 187 annotators across drone, crewed aviation, and satellite perspectives, this study reveals source-dependent revision patterns within a large-scale annotation workflow. While the evaluated data originates from post-disaster building damage assessments, the observed dynamics raise data-curation questions relevant to other multi-source remote sensing efforts under the limitations discussed in Section \ref{subsec:limitations}.  

The analysis has yielded two surprising findings. First, revision rates rose steeply from higher- to lower-resolution sources: the committee revised 25.27\% of initial crewed-aviation annotations and 36.95\% of satellite annotations, and the same ordering held at every observed workflow stage. Second, a single individual review reduced but did not resolve this disagreement, leaving 6.85\% (drone), 14.05\% (crewed), and 20.86\% (satellite) of labels to be revised by the consensus committee. These disparities pose risks of model bias and unreliability when such datasets train downstream computer vision models. To mitigate these risks, this evidence suggests that curation efforts move away from uniform review allocation and consider three strategies: (1) tailor labeling schemas to each imagery source, (2) prioritize consensus-based adjudication over individual review alone, and (3) preferentially target review capacity toward lower-resolution sources.

Future efforts will extend this work in three specific directions. First, revision rates among buildings annotated by multiple individuals, driven by overlapping image tiles, will be evaluated to measure baseline inter-annotator agreement prior to review intervention. Second, a fine-grained analysis of revision rates for specific class labels across sources will be conducted to identify which annotation targets are most susceptible to resolution degradation. Finally, high-performance computer vision models will be benchmarked directly against these human-curated labels to evaluate the operational boundaries of human-AI complementarity in remote sensing applications. 

\begin{acks}
This work is supported by the AI Research Institutes Program funded by the National Science Foundation under the
AI Institute for Societal Decision Making (NSF AI-SDM),
Award No. 2229881, and under “Datasets for Uncrewed
Aerial System (UAS) and Remote Responder Performance
from Hurricane Ian” Award No. 2306453. The authors thank the Center for Robot-Assisted Search and Rescue for access to the details of the drone data used in this work. The authors thank the US National Oceanic and Atmospheric Administration (NOAA) for releasing crewed aircraft imagery and Vantor (previously MAXAR) for releasing satellite imagery for these events. Further, the authors thank the annotators who participated in this work, specifically, the instructors and students at Winchester Thurston School, Bryan Collegiate High School, The Galveston Independent School District, Beaver Valley Intermediate Unit, Rudder High School, and Bryan Independent School District. Finally, the authors thank Jayesh Tripathi for his support during the review of the satellite and crewed aircraft data.
\end{acks}

\bibliographystyle{ACM-Reference-Format}
\bibliography{hcomp2026-14}

\appendix
\newpage
\section{Statistics}
\label{appendix:summary_stats}
This appendix section details the summary statistics for the building damage annotation task for each of the sources and each of the different data types. This information is shown in Table \ref{tab:summary_stats}.

\begin{table}[htbp!]
\centering
\begin{tabular}{|l|c|c|c|}
\hline
               & \textbf{Tiles} & \textbf{Buildings} & \textbf{Sub-Polygons} \\ \hline
Drone          & 19,609         & 20,041             & 40,761                \\ \hline
Crewed         & 13,121         & 20,695             & 45,582                \\ \hline
Satellite      & 10,493         & 33,392             & 66,835                \\ \hline
\textbf{Total} & 43,223         & 74,128             & 153,178               \\ \hline
\end{tabular}
\caption{A table summarizing the different quantities of items presented to annotators.}
\label{tab:summary_stats}
\end{table}
\section{Review Cards}
\label{appendix:review_cards}
\begin{figure}[htpb!]
    \centering
    \fbox{\includegraphics[width=0.75\columnwidth]{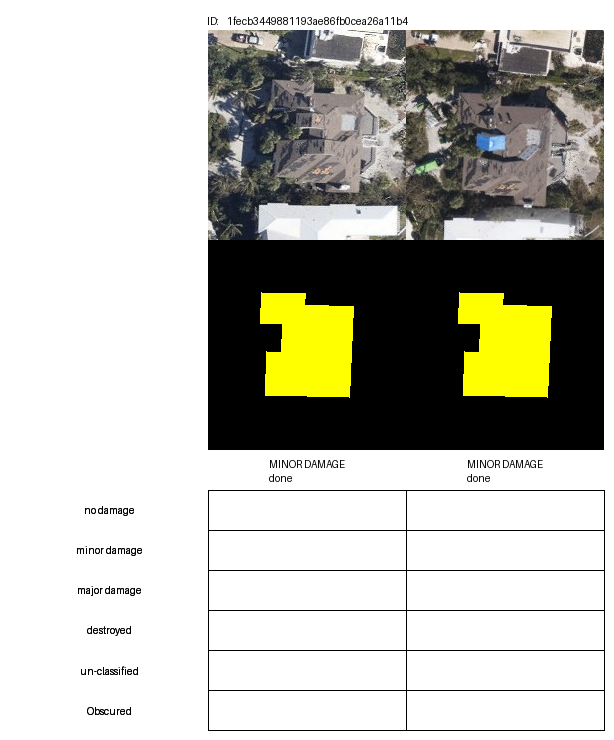}}
    \caption{An example blank Review Card that was used during the reviews of the Crewed and Satellite imagery. The committee would make corrections to building labels by coloring in the cell corresponding to the updated label for the imagery.}
    \Description{A figure which contains a review card for a single building. Two distinct views of this building are included in this card, with an associated polygon for each building. A 2-column, 6-row table is included below the views, showing the cells that can be marked to update the view's label.}
    \label{fig:review_cards}
\end{figure}

Review cards were visual representations of the multiple, parallel views of buildings that were used for the Final Committee Reviews for both Crewed and Satellite imagery. These cards captured the pixels from the orthomosaics that represent the building, the building polygon, the labels for those buildings and a grid corresponding to any updates that should be made to the labels for the building labels. An example of these review cards is shown in Figure \ref{fig:review_cards}. For crewed and satellite imagery, each review card presented the building's coincident views from both sources side by side.

\end{document}